\documentclass[10pt,twocolumn,letterpaper]{article}

\usepackage{cvpr}               %
\usepackage{graphicx}
\usepackage{amsmath}
\usepackage{amssymb}
\usepackage{booktabs}
\usepackage{rotating}
\usepackage{multirow}
\usepackage{pifont}
\newcommand{\cmark}{\ding{51}} %
\newcommand{\xmark}{\ding{55}} %

\usepackage[multiple]{footmisc}
\usepackage{bigfoot}

\DeclareNewFootnote{AAffil}[arabic]
\DeclareNewFootnote{ANote}[fnsymbol]

\usepackage[dvipsnames]{xcolor}
\usepackage[pagebackref,breaklinks]{hyperref}
\usepackage[accsupp]{axessibility}

\hypersetup{
    colorlinks=true,
    }

\usepackage[capitalize]{cleveref}
\crefname{section}{Sec.}{Secs.}
\Crefname{section}{Section}{Sections}
\Crefname{table}{Table}{Tables}
\crefname{table}{Tab.}{Tabs.}

\def\cvprPaperID{10529}  %
\def\confName{CVPR}
\def\confYear{2023}

\begin{document}

\title{Sampling is Matter: Point-guided 3D Human Mesh Reconstruction}

\author{Jeonghwan Kim$^{1}$\thanks{equal contribution}\quad Mi-Gyeong Gwon$^{1}$\footnotemark[1]\quad Hyunwoo Park$^{1}$\quad \\
Hyukmin Kwon$^{2}$\quad Gi-Mun Um$^{2}$\quad Wonjun Kim$^1$\thanks{corresponding author}\\
$^{1}$Konkuk University \quad $^{2}$Electronics and Telecommunications Research Institute\\
{\tt\small \{jhkim0759,kmk3942,pzls,wonjkim\}@konkuk.ac.kr \quad \{hmkwon,gmum\}@etri.re.kr}
}
\maketitle

\begin{abstract}
   This paper presents a simple yet powerful method for 3D human mesh reconstruction from a single RGB image.
   Most recently, the non-local interactions of the whole mesh vertices have been effectively estimated in the transformer while the relationship 
   between body parts also has begun to be handled via the graph model.
   Even though those approaches have shown the remarkable progress in 3D human mesh reconstruction, it is still difficult to directly infer the 
   relationship between features, which are encoded from the 2D input image, and 3D coordinates of each vertex.
   To resolve this problem, we propose to design a simple feature sampling scheme.
   The key idea is to sample features in the embedded space by following the guide of points, which are estimated as projection results of 3D mesh 
   vertices (i.e., ground truth). This helps the model to concentrate more on vertex-relevant features in the 2D space, thus leading to the 
   reconstruction of the natural human pose. Furthermore, we apply progressive attention masking to precisely estimate local interactions between 
   vertices even under severe occlusions. Experimental results on benchmark datasets show that the proposed method efficiently improves the 
   performance of 3D human mesh reconstruction.
   The code and model are publicly available at: \href{https://github.com/DCVL-3D/PointHMR_release}{\nolinkurl{https://github.com/DCVL-3D/PointHMR\_release}}.
\end{abstract}

\section{Introduction}
\label{sec:intro}

The goal of 3D human mesh reconstruction is to estimate 3D coordinates of points, which make up the human body surface. Since the high-quality 3D human model has been consistently required for various immersive applications, many studies have devoted considerable efforts to accurately reconstruct the 3D human mesh. In the early stage of this field, complex optimization techniques were adopted to generate the 3D human model based on the relationship between multiple scenes, which are acquired by using stereo or multiple-view camera systems. Recently, owing to the great success of deep learning, the problem of 3D human mesh reconstruction now can be resolved only with a single RGB image, thus the majority has begun to develop compact network architectures and efficient training strategies. Even though such deep learning-based approaches have shown the significant progress in 3D human mesh reconstruction, this task is still challenging due to severe occlusions by diverse human poses and depth ambiguities by the monocular setting.

\begin{figure}
\centering
\centerline{\includegraphics[width=1.0\columnwidth]{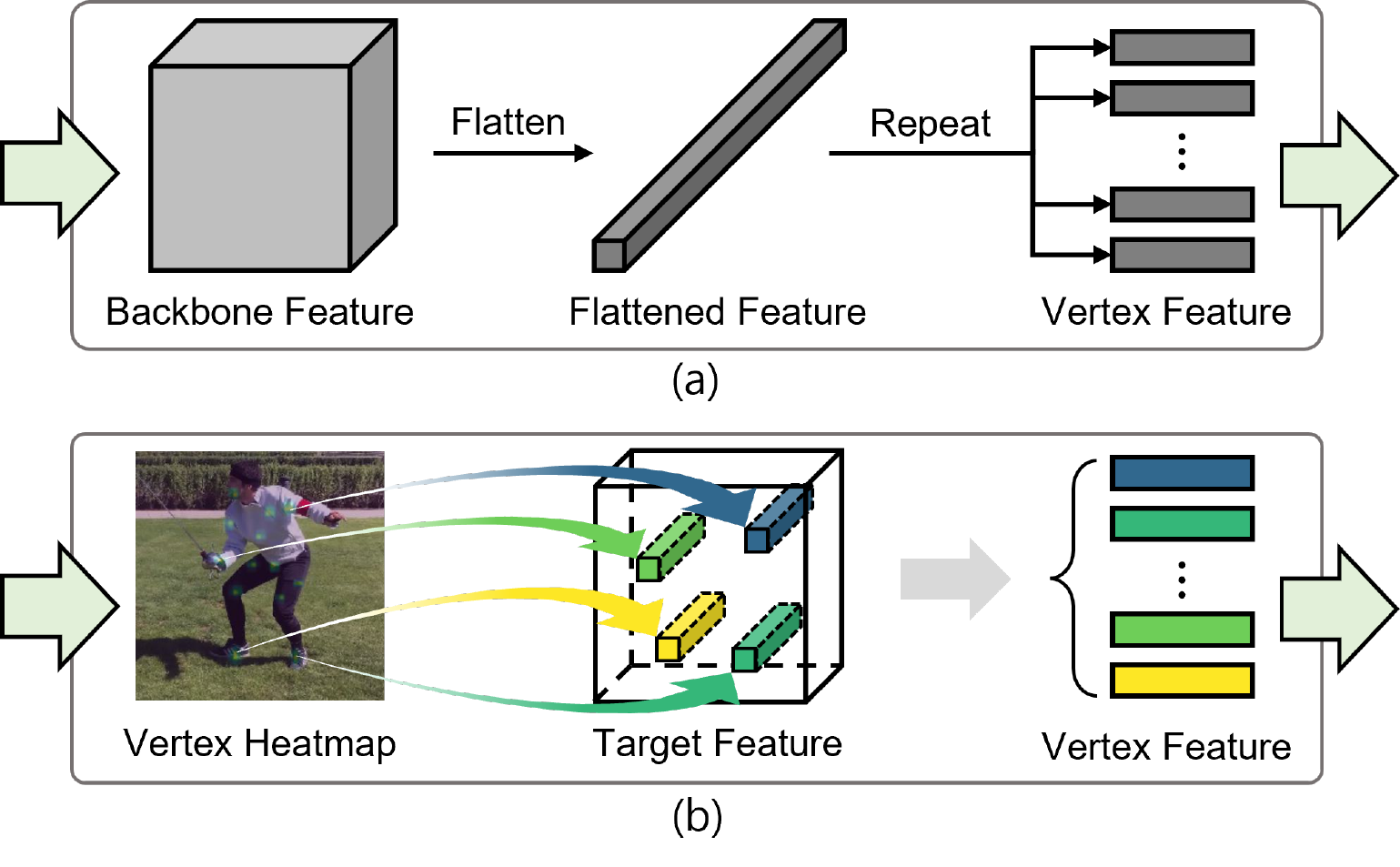}}
\caption{\label{fig:sampling}
(a) Traditional process of feature extraction for estimating 3D coordinates. (b) Vertex-relevant feature extraction process based on the proposed point-guided sampling method for estimating 3D coordinates.}

\end{figure}

Deep learning-based approaches can be divided into two main groups: model-based and model-free methods. In the former, most methods aim to estimate shape and pose parameters of the skinned multi-person linear (SMPL) model~\cite{Loper15}, which is capable of yielding the whole vertices via these two simple factors, thus most widely employed in this field. Traditional encoder-decoder architectures, which are mostly composed of stacked convolutional layers, are sufficient to conduct the regression for estimating those parameters. Despite their great performance, model-based methods have the obvious shortcoming, i.e., reconstruction results are limited to the pre-defined types of human body models. On the other hand, model-free methods have attempted to directly infer 3D coordinates of mesh vertices from input features without using any specific human body model. Compared to the model-based approach, which obtains the well-defined full mesh by adjusting shape and pose parameters, the model-free approach needs to estimate 3D coordinates of whole vertices directly from the network. Most methods in this category are based on the transformer to grasp non-local interactions between mesh vertices. The graph model (e.g., graph convolution) also has been utilized together to allow for body part relations in a local manner. One important advantage of the model-free approach is the flexibility to adapt to other applications, e.g., hand pose estimation, without significant changes of the data format and the training strategy. However, inferring the 3D coordinate from a single monocular image is still challenging due to lack of learning the correspondence between encoded features and spatial positions.

\begin{figure*}[t]
\vspace{-4mm}
\centerline{\includegraphics[width=1\textwidth]{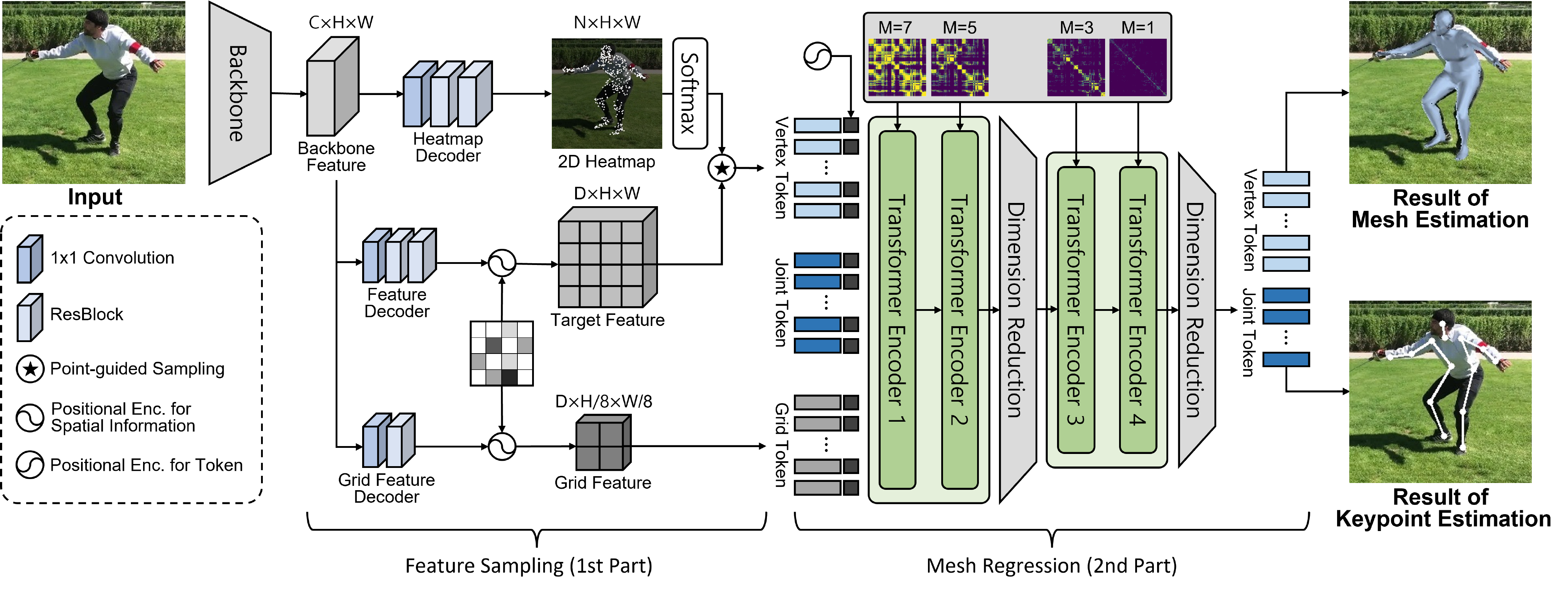}}
\vspace{-2mm}
\caption{\label{fig:overall_architecture} Overall architecture of the proposed method. The backbone feature is used to generate heatmap, target feature, and grid feature with respective decoders. Our sampling scheme utilizes heatmap and target feature to make the vertex token. Note that $N$ and $M$ denote the number of vertices and the distance threshold for defining the local connection in self-attention, respectively.
 }

\end{figure*}

In this paper, we propose a simple yet powerful method for 3D human mesh reconstruction. To this end, we conduct feature sampling at vertex-relevant points of the input image as shown in Fig.~\ref{fig:sampling}, which are estimated through the heatmap decoder trained by projection results of 3D mesh vertices (i.e., ground truth). These sampled features are subsequently fed into the transformer encoder as the form of the vertex token (see Fig.~\ref{fig:overall_architecture}). In a similar way of~\cite{Cho22}, we apply attention masking to the transformer encoder, however, the difference is that the local connection is defined with the range of multiple levels through the sequence of transformer encoders. This progressive attention masking helps the model understand local relations between vertices precisely even in occlusions. The main contribution of the proposed method can be summarized as follows:
\begin{itemize}
\item We propose to utilize the correspondence between encoded features and vertex positions, which are projected into the 2D space, via our point-guided feature sampling scheme. By explicitly indicating such vertex-relevant features to the transformer encoder, coordinates of the 3D human mesh are accurately estimated. 

\item Our progressive attention masking scheme helps the model efficiently deal with local vertex-to-vertex relations even under complicated poses and occlusions.
\end{itemize}

\section{Related Work}

In this Section, we give a brief review of the previous studies for 3D human mesh construction which have progressed in two different directions: model-based and model-free approaches.

\vspace{1mm}
\noindent\textbf{Model-based approaches.} As mentioned, most model-based approaches aim to estimate shape and pose parameters of the SMPL model~\cite{Loper15} for restoring the entire set of mesh vertices. In the beginning, several studies attempted to align 2D joint positions as well as body part segments, which are estimated by respective networks, with the ground truth projected from the 3D human mesh~\cite{Bogo16,Lassner17}. However, these methods require additional steps to estimate shape and pose parameters of the SMPL model. To cope with this limitation, Kanazawa {\it et al.}~\cite{Kanazawa18} proposed to regress such parameters directly from a single RGB image without using intermediate results, i.e., 2D joints and body part segments. Specifically, they designed a simple convolutional encoder with the adversarial loss to make the reconstructed mesh result be realistic. Inspired by the great potential of this simple regression scheme, many researchers have introduced various encoder-decoder architectures to estimate shape and pose parameters. Kolotouros {\it et al.}~\cite{Kolotouros19_ICCV} proposed to combine the end-to-end regression model with the optimization loop to strongly supervise the refinement process. Choutas {\it et al.}~\cite{Choutas20} attempted to directly regress body, face, and hands by exploiting the body-driven attention in the SMPL-X~\cite{Pavlakos19} format for generating the high-quality 3D human mesh. Biggs {\it et al.}~\cite{Biggs20} designed a multi-hypothesis neural network regressor based on the best-of-M loss, which makes the plausible human pose even under severely occluded environments. Zhang {\it et al.}~\cite{Zhang20} also focused on accurately restoring object-occluded human shape and pose by utilizing the partial UV map and the novel saliency map in the SMPL format. Most recently, Kocabas {\it et al.}~\cite{Kocabas21} devised the part attention module to guide the network to concentrate more on relevant body parts for inferring the given pose with SMPL parameters in a single input RGB image. Sun {\it et al.}~\cite{Sun21} estimated body center positions instead of segmenting human regions and extracted the parameter maps for the SMPL model at the corresponding positions. They further extend their algorithm by adopting the heat map of the bird-eye view to alleviate the depth ambiguity in monocular settings~\cite{Sun22}. On the other hand, the kinematic topology module has been embedded into the neural network architecture to consider the relationship between articulations of the human body~\cite{Li21}. Even though model-based approaches have shown the remarkable progress in 3D human mesh reconstruction, their performance is limited to pre-defined types of human body models, which are hardly extended to other applications.

\vspace{1mm}
\noindent\textbf{Model-free approaches.} Model-free approaches intend to directly restore the entire set of mesh vertices while relaxing the heavy reliance on the parameter space in model-based methods. As a pioneer, Kolotouros {\it et al.}~\cite{Kolotouros19_CVPR} proposed to estimate human mesh coordinates through the graph convolutional neural network (GraphCNN). To do this, they attached encoded features to nodes in the graph, which are mapped to 3D coordinates of the template mesh. From this perspective, Choi {\it et al.}~\cite{Choi20} also adopted GraphCNN to reconstruct 3D human meshes from 2D and 3D pose information in a coarse-to-fine manner. While these GraphCNN-based methods have a good ability to fully exploit the mesh topology, they are somewhat lacking in considering global interactions between joints and vertices. To cope with this limitation, the transformer has begun to be actively adopted for model-free approaches. Specifically, Lin {\it et al.}~\cite{Lin21} firstly introduced the transformer encoder, which simply takes joint and vertex queries as input tokens, to regress 3D coordinates from a single input RGB image. In particular, they further embedded GraphCNN into the transformer block to supplement local interactions, e.g., between-part relationships, for 3D human mesh reconstruction. Most recently, Cho {\it et al.}~\cite{Cho22} have disentangled image features and mesh queries by utilizing the transformer encoder-decoder architecture to alleviate the high complexity of interactions among input tokens.

Our method is also based on the transformer encoder with a simple sampling scheme, which gives a great help to focus on vertex-relevant features for inferring coordinates of the 3D human mesh. Technical details will be explained in the following Section.

\section{Proposed Method} 

The proposed method consists of two main parts. Specifically, vertex-relevant features are extracted based on our point-guided feature sampling in the first part while 3D coordinates are estimated through the sequence of transformer encoders with the proposed progressive attention masking scheme in the second part. The overall architecture of the proposed method is illustrated in Fig.~\ref{fig:overall_architecture}.

\begin{figure}
\centering
\centerline{\includegraphics[width=1.0\columnwidth]{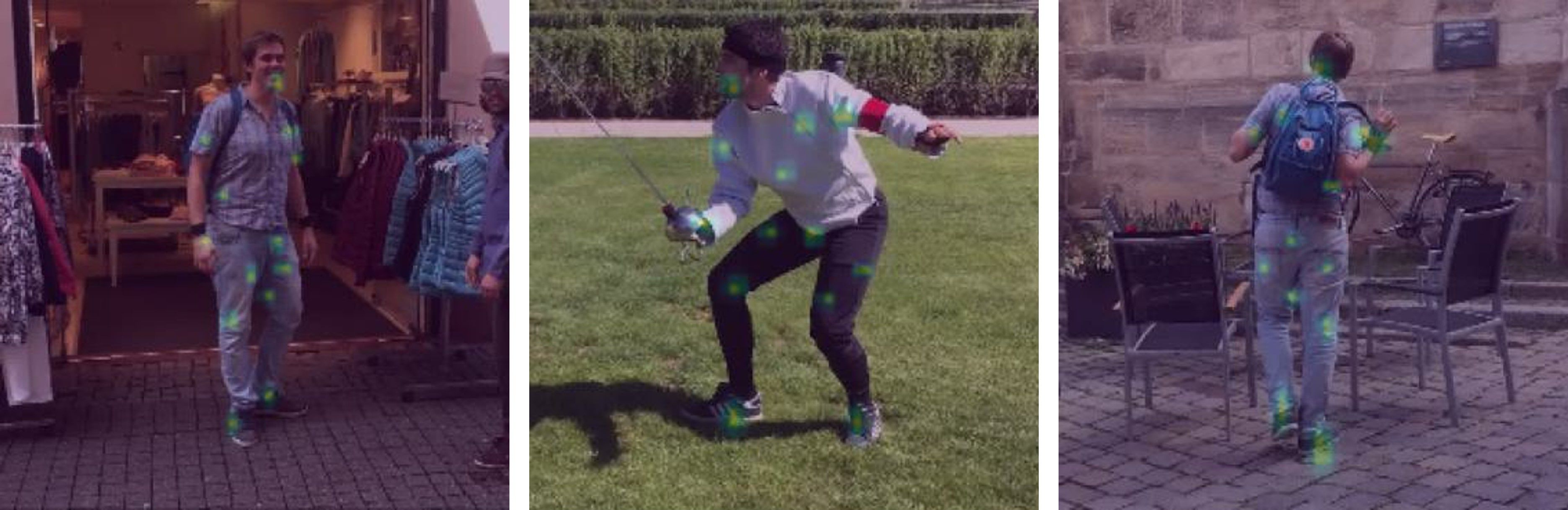}}
\vspace{-1mm}
\caption{\label{fig:heatmap} Several examples of predicted heatmaps on the 3DPW dataset. For better visibility, the activated regions of the heatmap corresponding to several selected vertices are represented in a single image with bright colors (best view in colors).}
\vspace{-3mm}
\end{figure}
\subsection{Point-guided Feature Sampling}

Since the direct transform from the color value to the 3D coordinate is still a difficult process due to heterogeneous modalities, we propose to use the intermediate guidance, i.e., features sampled at positions of vertices projected from 3D to 2D spaces. Specifically, we represent such projection results as the form of the heatmap, and sample features at activated positions in this heatmap.
The detailed process of our point-guided feature sampling is shown in the first part of Fig.~\ref{fig:overall_architecture}. Firstly, the backbone feature $X_b \in\mathbb{R}^{C\times H \times W}$ is encoded through the backbone network (HRNet~\cite{Wang21} in this work), where $C$, $H$, and $W$ denote the number of channels, height, and width, respectively.

$X_b$ is subsequently decoded into the heatmap, the target feature, and the grid feature by respective decoders as shown in Fig.~\ref{fig:overall_architecture}. After that, the vertex-relevant feature is sampled at each of $N$ vertices based on combination of the predicted heatmap and the target feature. 
Several examples of the predicted heatmap are shown in Fig.~\ref{fig:heatmap}. Note that we conducted element-wise multiplication of the predicted heatmap and the target feature to make our sampling process be differentiable. Moreover, we apply the positional encoding to the target feature in the same way of~\cite{Carion20,Cho22} for preserving the spatial information in the transformer encoder. Our point-guided feature sampling can be formulated as follows:
\begin{equation} \label{eq:feature_sampling}
\hat{V}_{i} = H_{i} \times F, \quad i = 1, 2, 3, ..., N,
\end{equation}
where $H_{i}$ and $F$ denote the predicted heatmap corresponding to the $i$-th vertex and the target feature, respectively. $N$ denotes the total number of vertices and is set to 431 as used in previous methods. $\hat{V}_i$ is the sampling result, which is finally flattened and fed into the transformer encoder with the grid feature in the second part of our proposed method.
It is noteworthy that the grid feature plays an important role to create the united body structure by aligning each point in an appropriate location.

\begin{figure}
\centering
\vspace{-4mm}
\centerline{\includegraphics[width=1.0\columnwidth]{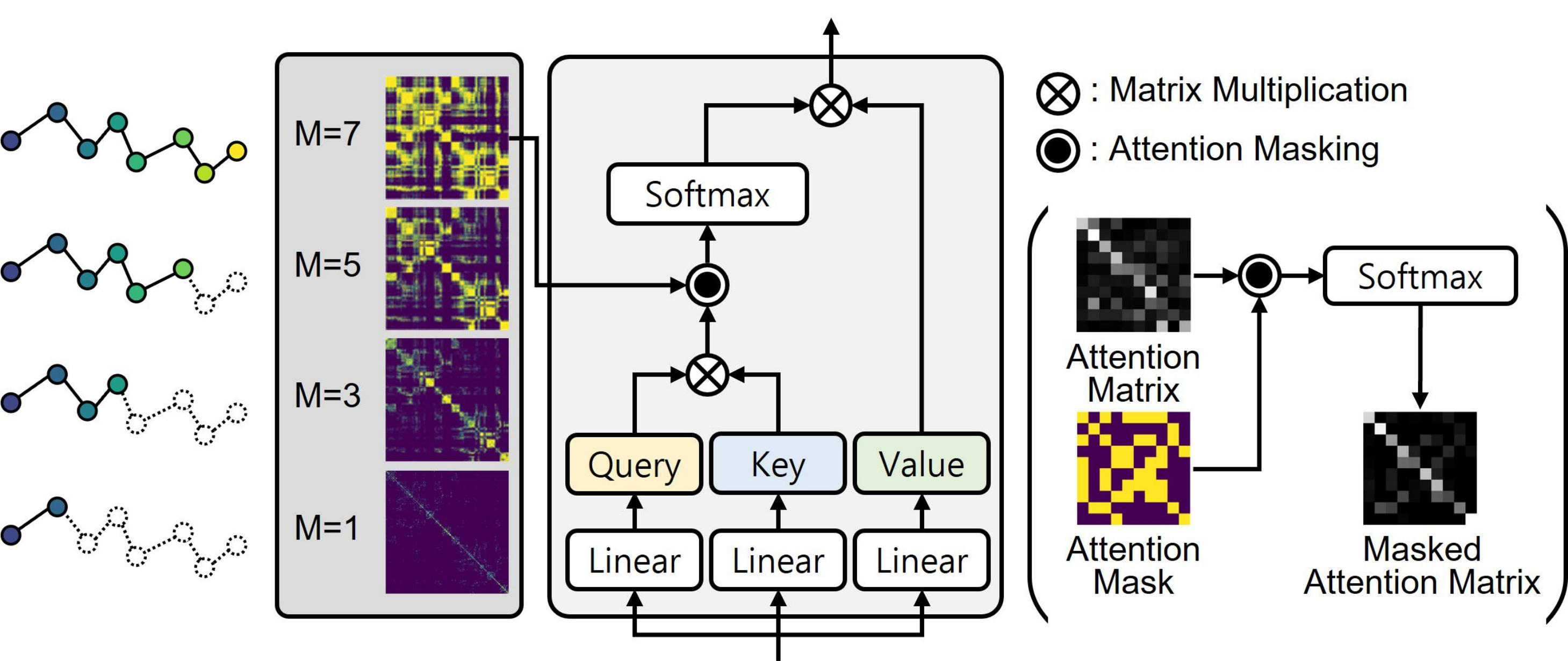}}
\caption{\label{fig:attention_masking} The detailed structure of the proposed progressive attention masking scheme. Note that this figure illustrates the masking process in the first transformer encoder, i.e., the case of $M=7$, shown in Fig.~\ref{fig:overall_architecture}.
}
\vspace{-1mm}
\end{figure}

\subsection{Transformer Encoder with Progressive Attention Masking}
\begin{figure}
\centering

\centerline{\includegraphics[width=1.0\columnwidth]{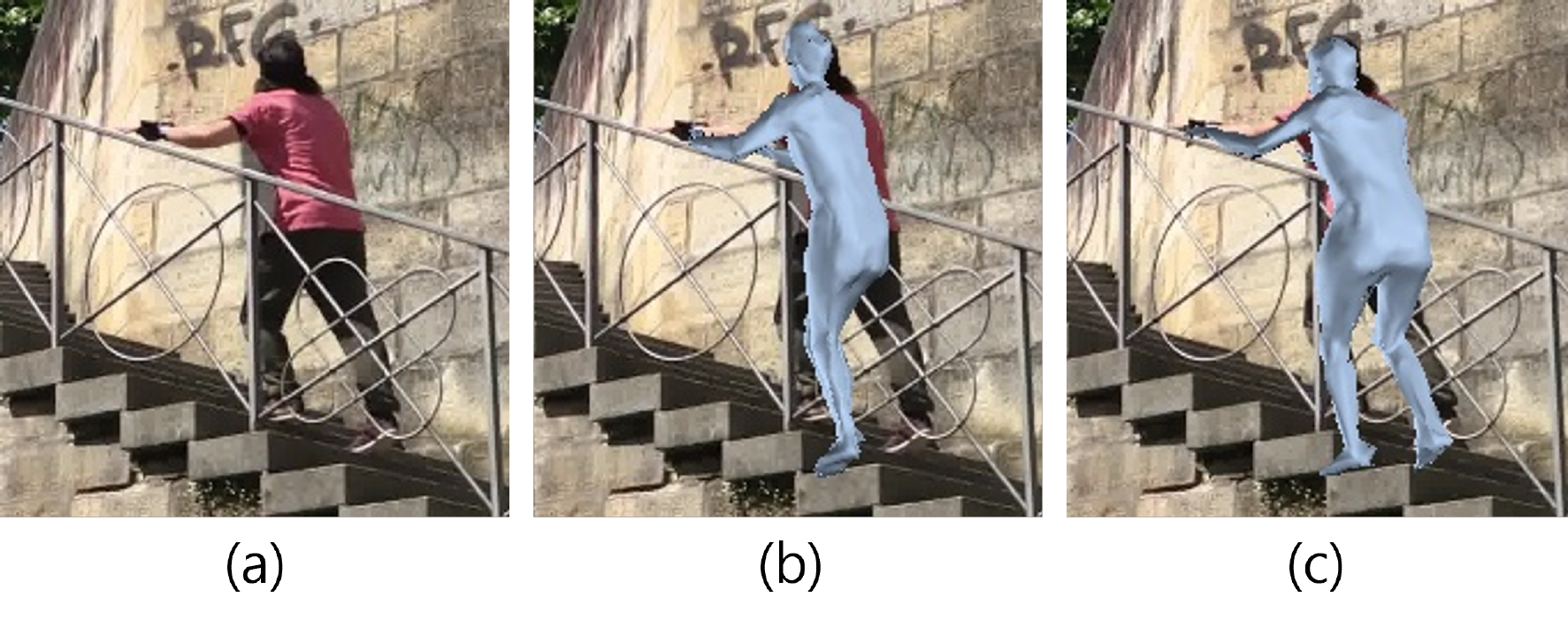}}
\vspace{-2.3mm}
\caption{\label{fig:vertex_connection_masking}(a) Input image. (b) Reconstruction result with previous attention masking~\cite{Cho22}. (c) Reconstruction result with progressive attention masking (proposed).}
\vspace{-1mm}
\end{figure}

In this subsection, we explain the sequence of transformer encoders in detail. In a similar way of~\cite{Lin21, Lin21_ICCV}, we design the transformer encoder with the dimension reduction layer as shown in the second part of Fig.~\ref{fig:overall_architecture}. 
Specifically, the transformer encoder takes the vertex token $\hat{V}\in\mathbb{R}^{N\times D}$, the joint token $\hat{J}\in\mathbb{R}^{K\times D}$, and the grid token $\hat{G}\in\mathbb{R}^{Z\times D}$ as inputs.
The joint token is a trainable parameter that is randomly initialized and optimized during the training phase whereas vertex and grid tokens are extracted from the first part of the proposed method.
Moreover, positional encodings are employed to give the identity by concatenating the trainable parameter to each token. The dimension of the encoded token is reduced by linear projection after each transformer block, which consists of two transformer encoders. To consider local vertex-vertex relations as well as non-adjacent interactions, we exclude far-distant connections between vertices to compute self-attention in the transformer encoder. In contrast to the previous method~\cite{Cho22}, we gradually decrease the range to define the local connection between vertices through the sequence of transformer encoders as illustrated in Fig.~\ref{fig:attention_masking}. This helps the model consider the local relationship between neighbor vertices in a progressive manner. Figure~\ref{fig:vertex_connection_masking} shows the effect of our progressive attention masking. As can be seen, the proposed method provides the reasonable result even in occlusions.

The outputs (i.e., vertex and joint tokens) of the last transformer encoder are finally projected into 3D coordinates via a linear layer. In the case of the vertex token, the upsampling algorithm introduced in \cite{ranjan18} is applied to expand sparse vertices $V'\in\mathbb{R}^{N \times3}$ into dense vertices $V\in\mathbb{R}^{\mathcal{N} \times3}$ as follows:
\begin{equation} \label{eq:feature_upsampling}
V = UV',
\end{equation}
where $U$ is the pre-defined upsampling matrix~\cite{ranjan18}. $\mathcal{N}$ is set to 6,890 (same as the vertex number of the SMPL model). 
The predicted 3D human mesh and 3D keypoints can be visualized as the rightmost images shown in Fig.~\ref{fig:overall_architecture}.

\begin{table*}[t]
\small
\centering
\begin{tabular}{c l c c c c c c}
\hline
\toprule
\multicolumn{2}{c}{\multirow{2}{*}[-0.7ex]{Methods}} & \multirow{2}{*}[-0.7ex]{Backbone} & \multicolumn{2}{c}{Human3.6M} & \multicolumn{3}{c}{3DPW}\\ \cmidrule(lr){4-5}\cmidrule(lr){6-8}
    & & & MPJPE($\downarrow$) & PA-MPJPE($\downarrow$) & MPJPE($\downarrow$) & PA-MPJPE($\downarrow$) & MPVPE ($\downarrow$)\\ \midrule

\multirow{10}{*}[-3.0ex]{\begin{turn}{90}model-based\end{turn}}  & HMR~\cite{Kanazawa18} & ResNet-50 & 88.0 & 56.8 & 130.0 & 81.3 & --\\ 
& SPIN~\cite{Kolotouros19_ICCV} & ResNet-50 & -- & 41.1 & 96.9 & 59.2 & 116.4\\
& ExPose~\cite{Choutas20} & ResNet-50 & -- & -- &  93.4 & 55.6 & -- \\ 
& VIBE~\cite{Kocabas20} & ResNet-50 & 65.6 & 41.4 &  82.9 &  51.9 & 99.1\\
& HybrIK~\cite{Li21} & ResNet-34 &54.4 & 34.5 & 80.0 & 48.8 & 94.5\\
& ROMP~\cite{Sun21} & HRNet-W32 & -- & -- & 76.7 & 47.3 & 93.4\\
& PARE~\cite{Kocabas21} & HRNet-W32 & -- & -- & 74.5 & 46.5 & 88.6\\
& MAED~\cite{Wan21} & ResNet-50 & 56.4 & 38.7 & 79.1 & 45.7 & 92.6  \\
& PyMAF~\cite{Zhang21} & ResNet-50 & 57.7 & 40.5 & 92.8 & 58.9 & 110.1  \\
& BEV*~\cite{Sun22} & HRNet-W32 & -- & -- & 78.5 & 46.9 & 92.3\\
& OCHMR*~\cite{Khirodkar22} & ResNet-50 / HRNet-W32 & -- & -- & 89.7 & 58.3 & 107.1\\
& 3DCrowdNet*~\cite{Choi22} & ResNet-50 & -- & -- & 81.7 & 51.5 & 98.3\\
\midrule
\multirow{7}{*}[1.2ex]{\begin{turn}{90}model-free\end{turn}} & GraphCMR~\cite{Kolotouros19_CVPR} & ResNet-50 & -- & 50.1 & -- & 70.2 & --\\
&I2LMeshNet~\cite{Moon2020} & ResNet-50 & 55.7 & 41.7 & 93.2 & 57.7 & 110.1\\
&Pose2Mesh~\cite{Choi20} & HRNet-W48 & 64.9 & 47.0 & 89.5 & 56.3 & 105.3\\
&METRO~\cite{Lin21} & HRNet-W64 & 54.0 & 36.7 & 77.1 & 47.9 & 88.2\\
&MeshGraphormer~\cite{Lin21_ICCV} & HRNet-W64 & 51.2 & 34.5 & 74.7 & 45.6 & 87.7\\
&FastMETRO~\cite{Cho22} & HRNet-W64 & 52.2 & 33.7 & \textbf{73.5} & \textbf{44.6} & \textbf{84.1}\\
\hline
&Ours & HRNet-W32 & \textbf{48.3} & \textbf{32.9} & 73.9 & 44.9 & 85.5\\ \bottomrule
\hline
\end{tabular}
\caption{\label{table:performance_total} Performance comparisons of 3D human mesh reconstruction based on Human3.6M and 3DPW datasets. The proposed method achieves the best performance in the Human3.6M dataset while still showing the competitive performance in the 3DPW dataset (best results are shown in bold). Note that * denotes the performance without fine-tuning on the 3DPW dataset.}

\end{table*}

\subsection{Loss Function}
The proposed method is trained based on four types of loss functions, i.e., vertex loss $\mathcal{L}_{v}$, 3D joint loss $\mathcal{L}_{j3d}$, 2D joint loss $\mathcal{L}_{j2d}$, and heatmap loss $\mathcal{L}_{h}$.
The first three loss functions are used for estimating positions of vertices and joints as introduced in previous works~\cite{Lin21, Lin21_ICCV, Cho22} while the last one guides the network to find positions of projected vertices.
First of all, $L_1$ loss is adopted to compute the difference between positions of the predicted vertex $V$ and the corresponding ground truth $\tilde{V}$ as follows:
\begin{equation} \label{eq:loss_vertex}
 \mathcal{L}_{v} = \frac{1}{N}\sum_{i=1}^{N}{\Vert \tilde{V}^i-V^i \Vert_1},
\end{equation}
where $N$ denotes the total number of vertices.
For the 3D joint loss, we compute the distance between the estimated joint position $J_{3d}$ and the corresponding ground truth $\tilde{J}_{3d}$ in the 3D space. Note that the joint position, which is regressed from the predicted vertices, i.e., $\bar{J}_{3d}$, is also used for the loss computation as follows:
\begin{equation} \label{eq:loss_3D}
 \mathcal{L}_{j3d} = \frac{1}{K}\sum_{i=1}^{K}{\Vert \tilde{J}^i_{3d}-J^i_{3d} \Vert_2 + \Vert \tilde{J}^i_{3d}-\bar{J}^{i}_{3d} \Vert_2},
\end{equation}
where $K$ denotes the total number of joints.
Similarly, the 2D joint loss is calculated as well. To do this, $J_{3d}$ and $\bar{J}_{3d}$ are projected onto the 2D space and the corresponding results are represented as $J_{2d}$ and $\bar{J}_{2d}$, respectively.
Based on such projected results, the 2D joint loss is formulated as $L_2$ loss in the same way of the 3D joint loss as follows:
\begin{equation} \label{eq:loss_2D}
 \mathcal{L}_{j2d} = \frac{1}{K}\sum_{i=1}^{K}{\Vert \tilde{J}^i_{2d}-J^i_{2d} \Vert_2 + \Vert \tilde{J}^i_{2d}-\bar{J}^i_{2d} \Vert_2},
\end{equation}
where $\tilde{J}_{2d}$ is the ground truth of the 2D joint position.
On the other hand, our heatmap loss $\mathcal{L}_{h}$ consists of the binary cross entropy loss and the dice loss. Specifically, we adopt the binary cross entropy loss to determine whether the activated position is matched to the point projected from the vertex or not as follows:
\begin{equation}
\label{eq:loss_heatmap_bce}
   \mathcal{L}_{bce}  = -\frac{1}{N}\sum_{i=1}^{N}\tilde{H}^{i}\log{H^{i}} + (1-\tilde{H}^{i})\log{(1-{H}^{i})},
\end{equation}
where $H$ and $\tilde{H}$ denote the predicted heatmap and the corresponding ground truth, respectively. The ground truth is represented as the binary map where the position of the projected vertex is assigned 1, otherwise 0.
To alleviate the data-imbalanced problem, i.e., the projected point exists on only a single pixel in the ground truth image, the dice loss~\cite{Milletari16} is also employed as follows:
\begin{equation}
\label{eq:loss_heatmap_dice}
   \mathcal{L}_{dice} = \frac{1}{N}\sum_{i=1}^{N} 1- \frac{2 \times (\tilde{H}^{i} \times H^{i})}{\tilde{H}^{i} + H^{i}}.
\end{equation}
Since the dice loss mainly focuses on the overlapped area, the data-imbalanced problem can be efficiently alleviated. 
By using the combination of these loss functions, the proposed network successfully learns to reconstruct the 3D human mesh as follows: 
\begin{equation} 
\begin{aligned}
\label{eq:loss_total}
 & \mathcal{L}_{total} = w_{v}\mathcal{L}_{v} + w_{j3d}\mathcal{L}_{j3d} + w_{j2d}\mathcal{L}_{j2d} \\
 & \qquad \quad \enspace  + w_{bce}\mathcal{L}_{bce} + w_{dice}\mathcal{L}_{dice},
\end{aligned}
\end{equation}
where $w_{v}$, $w_{j3d}$, $w_{j2d}$, $w_{bce}$ and $w_{dice}$ are the balancing factor for each loss term, which are set to 0.01, 0.1, 0.01, 1.0, and 0.001, respectively.

\begin{figure*}[t]
\vspace{-4mm}
\centerline{\includegraphics[width=1\textwidth]{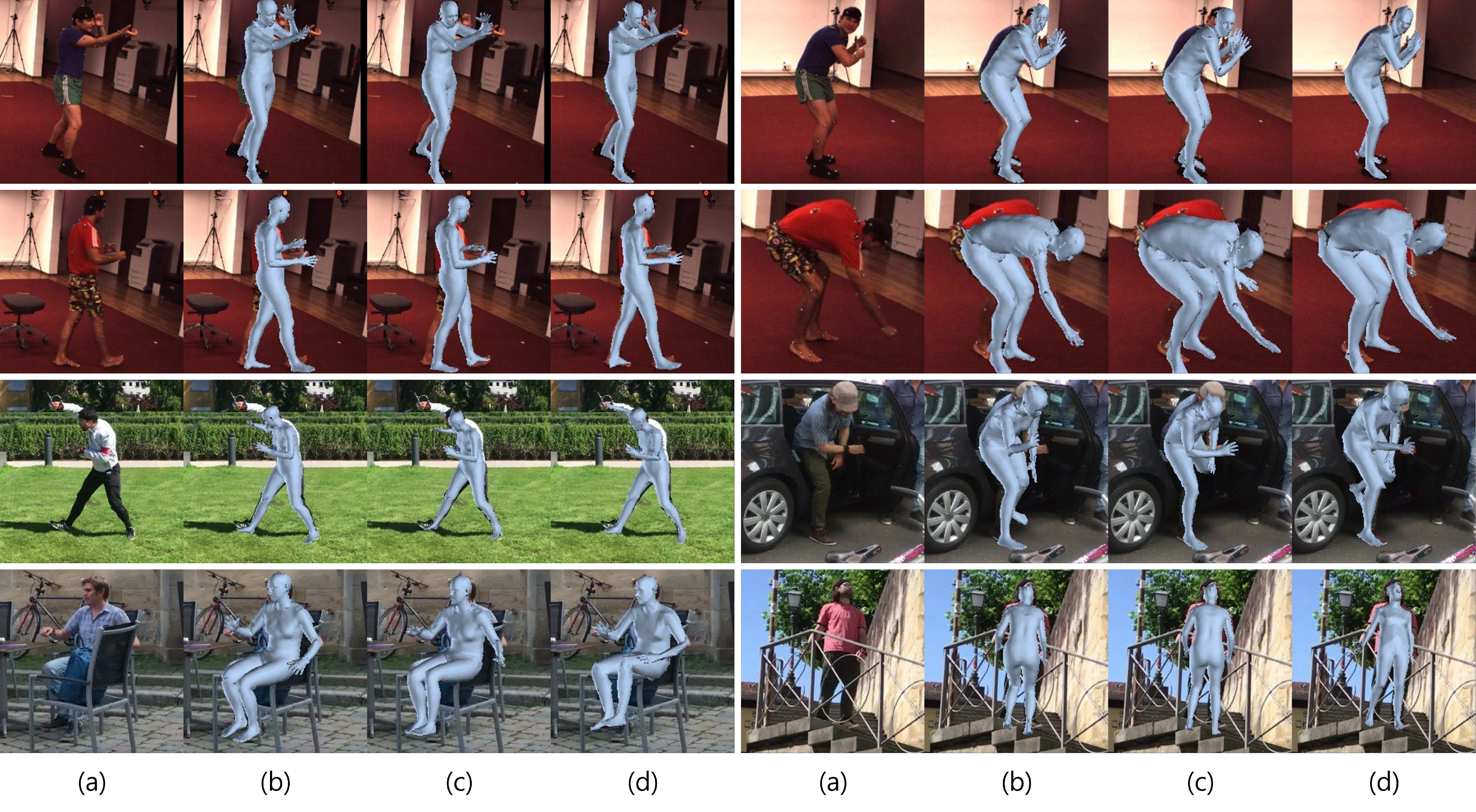}}
\caption{\label{fig:comparison}
Results of 3D human mesh reconstruction on Human3.6M~\cite{Ionescu13} (top-two rows) and 3DPW~\cite{Von2018} (bottom-two rows) datasets. (a) Input images. (b) Results by METRO~\cite{Lin21}. (c) Results by Mesh Graphormer~\cite{Lin21_ICCV}. (d) Results by the proposed method.}
\end{figure*}

\section{Experimental Results}
\subsection{Training}
All the experiments are implemented on the PyTorch framework~\cite{Paszke17} with an Intel E5-1650@3.60GHz CPU and two NVIDIA RTX A6000 GPUs. To train all the parameters of our model, the Adam optimizer~\cite{Kingma2015} is adopted where the momentum factors are set to 0.9 and 0.999, respectively. The proposed network is trained for 50 epochs with a batch size of 64 per GPU. The learning rate is firstly set to $1\times{10^{-4}}$ and reduced to $1\times{10^{-5}}$ at the half of the learning time. For each input image, the area including the human is cropped and resized to the resolution of $224\times224$ pixels before training.

\subsection{Datasets and Evaluation Metrics}
\vspace{1mm}
\noindent\textbf{Datasets.} For the performance evaluation of the proposed method, two representative benchmarks, i.e., Human3.6M~\cite{Ionescu13} and 3DPW~\cite{Von2018}, are employed. Specifically, the proposed method is trained based on five datasets, i.e., Human3.6M~\cite{Ionescu13}, MuCo-3DHP~\cite{Mehta18}, UP-3D~\cite{Lassner17}, COCO~\cite{Lin14}, and MPII~\cite{Andriluka14}, by following previous approaches, and tested with the P2 protocol in the Human3.6M dataset. Since the ground truth of the 3D human mesh is unavailable in the Human3.6M dataset, we use the pseudo mesh label generated by SMPLify-X~\cite{Pavlakos19} as introduced in~\cite{Choi20, Moon2020, Lin21, Lin21_ICCV, Cho22}. For the test on the 3DPW dataset, we fine-tune the proposed method by using the training set of the 3DPW dataset.

\vspace{1mm}
\noindent\textbf{Evaluation metrics.} For the quantitative evaluation, we use three metrics, i.e., mean per joint position error (MPJPE)~\cite{Ionescu13}, Procrustes-aligned mean per joint position error (PA-MPJPE)~\cite{Zhou18}, and mean per vertex position error (MPVPE)~\cite{Pavlakos18}, which have been widely adopted for the performance comparison in this field. Specifically, MPJPE measures the average value of the Euclidean distance between each estimated 3D joint and the corresponding ground truth. PA-MPJPE indicates MPJPE in which the estimated human body is aligned in terms of rotation and scaling using the Procrustes analysis. On the other hand, MPVPE is a metric for computing the Euclidean distance between coordinates of the predicted vertex and the corresponding ground truth.

\begin{figure*}[t]
\vspace{-4mm}

\centerline{\includegraphics[width=1\textwidth]{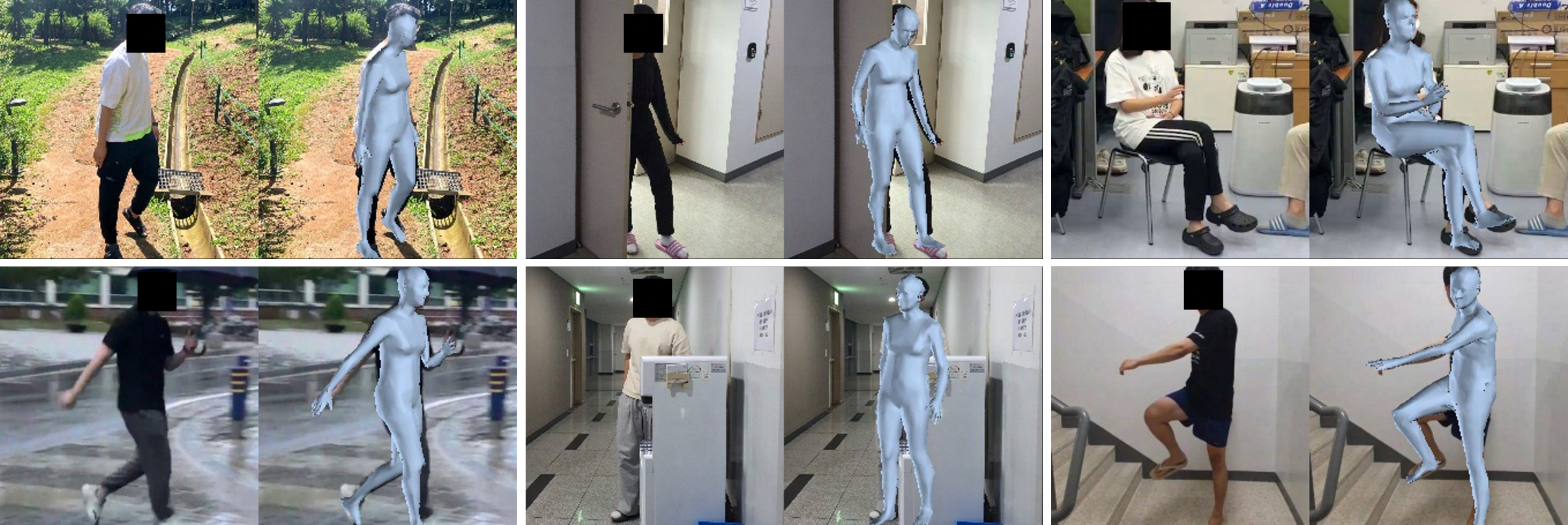}}
\caption{\label{fig:more_examples}
More results of 3D human mesh reconstruction for our samples, which are acquired by the smartphone. Odd columns: input images. Even columns: our results. Note that the proposed method provides reliable results of 3D human mesh reconstruction for various pictures of our daily life even with severe occlusions.}
\end{figure*}

\subsection{Performance Evaluation}
\noindent\textbf{Quantitative evaluation.} To demonstrate the efficiency and robustness of the proposed method, we compare ours with representative methods for 3D human mesh reconstruction, i.e., HMR~\cite{Kanazawa18}, SPIN~\cite{Kolotouros19_ICCV}, ExPose~\cite{Choutas20}, VIBE~\cite{Kocabas20}, HybrIK~\cite{Li21}, ROMP~\cite{Sun21}, PARE~\cite{Kocabas21}, MAED~\cite{Wan21}, PyMAF~\cite{Zhang21}, BEV~\cite{Sun22}, OCHMR~\cite{Khirodkar22}, 3DCrowdNet~\cite{Choi22}, GraphCMR~\cite{Kolotouros19_CVPR}, I2LMeshNet~\cite{Moon2020}, Pose2Mesh~\cite{Choi20}, METRO~\cite{Lin21}, Mesh Graphormer~\cite{Lin21_ICCV}, and FastMETRO~\cite{Cho22}. The result of the performance comparison is shown in Table~\ref{table:performance_total}. As can be seen, the proposed method achieves 48.3 MPJPE and 32.9 PA-MPJPE on the Human3.6M dataset, which outperforms the state-of-the-art methods with the meaningful performance gain. Even though the performance of the proposed method is slightly dropped compared to the best one (i.e., FastMETRO) in the 3DPW dataset, our method still shows the competitive performance with state-of-the-art methods. Specifically, model-free methods have shown the reliable performance without using the well-defined human model in recent days. This is because their reconstruction results are not limited to the small set of pre-defined human models, thus show more appropriate human meshes for a given image compared to model-based methods. In particular, transformer-based architectures, e.g., Mesh Graphormer and FastMETRO, significantly improves the performance of 3D human mesh reconstruction. Nevertheless, recent model-based methods still show the competitive performance and work robust to occlusion cases, e.g., BEV, OCHMR, and 3DCrowdNet. It is noteworthy that the proposed method notably improves the performance by our point-guided feature sampling scheme without specially designing the decoder architecture like~\cite{Cho22}. Moreover, the proposed method provides the reliable performance with the relatively small-sized backbone (e.g., HRNet-W32).

\vspace{1mm}
\noindent\textbf{Qualitative evaluation.} Several results of 3D human mesh reconstructions for Human3.6M and 3DPW datasets are shown in Fig.~\ref{fig:comparison}. Note that our results are reconstructed based on the upsampling matrix as used in~\cite{Cho22} while other two methods, i.e., METRO~\cite{Lin21} and Mesh Graphormer~\cite{Lin21_ICCV}, utilized two linear layers to upsample coarse vertex points. We can see that the proposed method successfully estimates 3D human poses under various situations including real-world scenarios in outdoor scenes as well as the controlled environment. Specifically, previous methods have somewhat difficulties to estimate unusual poses, e.g., overlapping arms and bending, whereas the proposed method provides the well-fit 3D model for a given image. Moreover, the proposed method has a good ability to reconstruct 3D human meshes in occlusions due to the progressive attention masking scheme as shown in examples of the last row of Fig.~\ref{fig:comparison}. In particular, previous methods fail to infer the global orientation of the human body due to the fence in front of the target person, which leads to the significant performance drop for 3D human mesh reconstruction. In contrast, the proposed method shows the reliable performance with various occlusions (see third and fourth rows in Fig.~\ref{fig:comparison}). More examples for 3D human mesh reconstruction by the proposed method are shown in Fig.~\ref{fig:more_examples}. Note that input images are acquired by the smartphone. As can be seen, the proposed method performs well for various pictures of our daily life even with severe occlusions. Therefore, it is thought that the proposed method paves the way for 3D human mesh reconstruction under various environments.

\begin{table}[t]
\small
\begin{center}
\begin{tabular}{c|c|c|c}
\hline
Point-guided  &  Progressive & \multirow{2}{*}[-0.3ex]{MPJPE} & \multirow{2}{*}[-0.3ex]{PA-MPJPE} \\
feature sampling & attention masking & \\
\hline
\hline
\xmark & \xmark & 63.2 & 39.9 \\
\xmark & \cmark & 61.2 & 40.8 \\
\cmark & \xmark & 50.9 & 33.3 \\
\cmark & \cmark & \textbf{48.3} & \textbf{32.9}\\
\hline
\end{tabular}
\caption{\label{table:ablation}Performance analysis of the proposed method according to changes in the network architecture based on the Human3.6M dataset (best results are shown in bold).}
\vspace{2mm}
\end{center}
\end{table}

\begin{table}[t]
\small
\begin{center}
\begin{tabular}{c|c|c}
\hline
Methods & {MPJPE} & {PA-MPJPE} \\
\hline
\hline
Without attention masking & 50.9 & 33.3 \\
Single attention masking & 49.1 & 33.5 \\
Progressive attention masking (ours) & \textbf{48.3} & \textbf{32.9} \\
\hline
\end{tabular}
\caption{\label{table:ablation2}Performance analysis of the proposed method according to the attention masking scheme based on the Human3.6M dataset (best results are shown in bold). Note that $M=1$ is used for single attention masking.}

\end{center}

\end{table}

\begin{figure*}[t]
\vspace{-4mm}

\centerline{\includegraphics[width=1\textwidth]{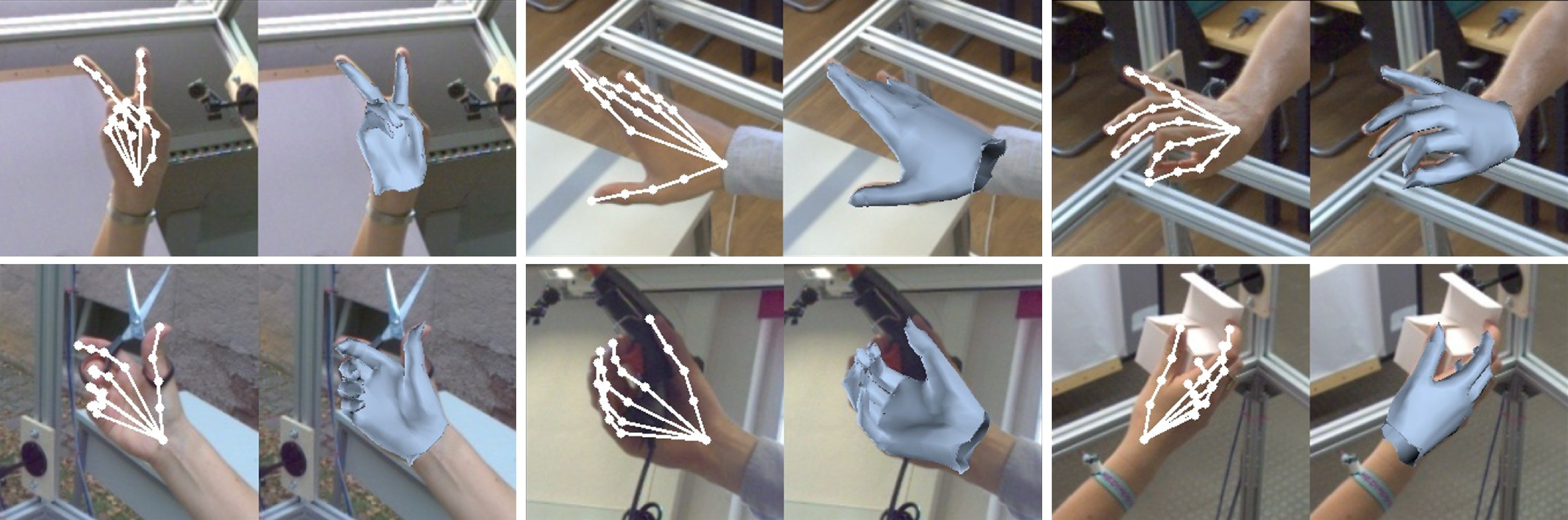}}
\caption{\label{fig:hand_results}
Several reconstruction results (3D joints (left) and 3D hand meshes (right)) by the proposed method for the FreiHAND~\cite{Zimmermann19} dataset. Note that the proposed method works robust to self-occlusions frequently occurring by complicated hand poses.}
\end{figure*}

\subsection{Ablation Studies}
In this subsection, we first demonstrate the comparative experimental results by changing the components of the proposed method based on the Human3.6M dataset. Table~\ref{table:ablation} shows the contribution of such components. As can be seen, the performance of 3D human mesh reconstruction is significantly improved with our point-guided feature sampling scheme (MPJPE: $63.2\rightarrow 50.9$, PA-MPJPE: $39.9\rightarrow 33.3$). From this analysis, we can see that feature sampling at vertex-relevant positions even in the 2D space is effective for improving the performance of 3D human mesh reconstruction. In what follows, we also check the effect of our progressive attention masking and the corresponding result is shown in Table~\ref{table:ablation2}. Note that our point-guided feature sampling is applied for this experiment. By comparing ours with baseline (i.e., without masking) and single masking attention scheme~\cite{Cho22}, we can see that the progressive restriction strategy in defining the local connection is also helpful for improving the performance of 3D human mesh reconstruction. This tells us that the combination of our contributions makes the model be robust to complicated real-world environments.

\subsection{Generalization Ability}
In contrast to the model-based approach, the proposed method can be easily applied to other applications. To show the generalization ability of the proposed method, we conduct 3D hand mesh reconstruction based on the FreiHAND~\cite{Zimmermann19} dataset by changing the number of input tokens for the sequence of transformer encoders. Note that the number of vertices to be projected is set to 195 and those are upsampled to 778 via the same upsampling matrix used for 3D human mesh reconstruction. Several reconstruction results by the proposed method are shown in Fig.~\ref{fig:hand_results}. We can see that the proposed method provides the reliable reconstruction results (i.e., 3D joints and 3D hand meshes) for various hand poses. In particular, self-occlusions by complicated relations between adjacent fingers are successfully handled in the proposed method. The result of the quantitative evaluation is also shown in Table~\ref{table:hand_ablation}. As can be seen, the proposed method shows the competitive performance on the FreiHAND dataset.

\begin{table}[t]
\footnotesize
\begin{center}
\begin{tabular}{c|@{\;\,}c@{\;\,}|@{\;}c@{\;}|@{\;\,}c@{\;\,}|@{\;}c@{\;}}
\hline
Methods & {PA-MPVPE} & {PA-MPJPE} & F@5mm & F@15mm \\
\hline
\hline
Hasson \textit{et al.}~\cite{Hasson19} & 13.2 & -- & 0.436 & 0.908 \\
Boukhayma \textit{et al.}~\cite{Boukhayma19} & 13.0 & -- & 0.435 &  0.898 \\
FreiHAND~\cite{Zimmermann19} & 10.7 & -- & 0.529 & 0.935\\
I2LMeshNet~\cite{Moon2020} & 7.6 & 7.4 & 0.681 & 0.973 \\
Pose2Mesh~\cite{Choi20} & 7.8 & 7.7 & 0.674 &  0.969 \\
METRO~\cite{Lin21} & 6.7 & 6.8 & 0.717 & 0.981 \\
FastMETRO~\cite{Cho22} & -- & 6.5 & -- & 0.982 \\
\hline
Ours & \textbf{6.6} & \textbf{6.1} &\textbf{ 0.720} &\textbf{ 0.984} \\
\hline
\end{tabular}
\caption{\label{table:hand_ablation}Performance comparisons of 3D hand mesh reconstruction based on the FreiHAND dataset (best results are shown in bold).}
\end{center}
\vspace{-5mm}
\end{table}

\section{Conclusion}
In this paper, we present a simple yet powerful method for 3D human mesh reconstruction from a single RGB image. The key idea of the proposed method is to alleviate heterogeneous modalities between input (i.e., color) and output (i.e., coordinate) by considering the correspondence of encoded features and 2D points projected from 3D vertices. Our point-guided feature sampling scheme is to sample vertex-relevant features based on the combination of the heatmap and encoded features. In addition, the proposed progressive attention masking scheme makes the model to be robust to occlusions by considering local connections of different levels through the sequence of transformer encoders. Experimental results on benchmark datasets show that the proposed method performs reliably for various real-world environments. 

\noindent\textbf{Acknowledgments.} This work was supported by Institute of Information \& communications Technology Planning \& Evaluation(IITP) grant funded by the Korea government(MSIT) (2021-0-02084, eXtended Reality and Volumetric media generation and transmission technology for immersive experience sharing in noncontact environment with a Korea-EU international cooperative research).

{\small
\bibliographystyle{ieee_fullname}
\bibliography{egbib}
}

\end{document}